\documentclass{article}
\usepackage[utf8]{inputenc}
\usepackage{spconf,amsmath,graphicx}
\usepackage{tabularx}
\usepackage{tabu}
\usepackage{natbib}
\usepackage[colorlinks=true,linkcolor=blue,citecolor=blue]{hyperref}

\usepackage{comment}
\usepackage{tikz}
\usepackage{forest}
\usetikzlibrary{trees,positioning,shapes,shadows,arrows.meta}

\title{Efficiency-oriented approaches for \\ self-supervised speech representation learning}

\name{Luis Lugo and Valentin Vielzeuf} 
\address{luiseduardo.lugomartinez@orange.com, valentin.vielzeuf@orange.com \\ Orange, Cesson-Sévigné, France}

\begin{document}

\maketitle
\begin{abstract}
Self-supervised learning enables the training of large neural models without the need for large, labeled datasets. It has been generating breakthroughs in several fields, including computer vision, natural language processing, biology, and speech. In particular, the state-of-the-art in several speech processing applications, such as automatic speech recognition or speaker identification, are models where the latent representation is learned using self-supervised approaches. Several configurations exist in self-supervised learning for speech, including contrastive, predictive, and multilingual approaches. There is, however, a crucial limitation in most existing approaches: their high computational costs. These costs limit the deployment of models, the size of the training dataset, and the number of research groups that can afford research with large self-supervised models. Likewise, we should consider the environmental costs that high energy consumption implies. Efforts in this direction comprise optimization of existing models, neural architecture efficiency, improvements in finetuning for speech processing tasks, and data efficiency. But despite current efforts, more work could be done to address high computational costs in self-supervised representation learning.
\end{abstract}
\section{Introduction}
\label{sec:intro}

Labeling data for speech recognition models is expensive, a limitation that hinders research and industrial applications in speech processing \citep{hsu2021hubert,mohamed2022self}. Moreover, low-resource languages and dialects that have no written resources create limitations for supervised training of deep learning architectures. These limitations motivate the need for self-supervised models to learn latent speech representations \citep{hsu2021hubert,zhang2020pushing}. 

Self-supervised approaches for speech representation learning can be based on consistency or self-training \citep{zhang2020pushing}. In self-training, a teacher model generates labels to train a student model. The self-training process is iterative, using several iterations to get the final model. Combined SSL, for instance,  uses four iterations to train a state-of-the-art speech representation model \citep{zhang2020pushing}. There is also the possibility to train models in a semi-supervised way, combining labeled and unlabeled data in the training dataset. In such cases, methods use a supervised loss for the labeled data, combining it with an unsupervised objective or pseudo labels for the unlabeled data \citep{huang2022spiral}. Noisy student from computer vision is an example \citep{xie2020self}.

On the other hand, consistency defines a task and trains the model on unlabeled data to learn a latent representation \citep{zhang2020pushing}. For example, bootstrap your own latent (BYOL) trains a model comprising online and target networks, which interact and learn from each other using augmented samples of input data \citep{grill2020bootstrap}. Once models learn a latent representation, they can be fine-tuned for a downstream task -- automatic speech recognition or speaker identification, for example  -- in a supervised way \citep{chen2020simple,huang2022spiral,zhang2020pushing}.

Currently, self-supervised models generate state-of-the-art results when learning latent representations, and their results in speech processing are competitive \citep{parcollet2023efficiency}. Downstream tasks in speech processing strongly benefit from latent representations \citep{mohamed2022self}. Those tasks include phoneme recognition (PR), keyword spotting (KS), intent classification (IC), speaker identification (SID), emotion recognition (ER), automatic speech recognition (ASR), query by example spoken term detection (QbE), slot filling (SF), automatic speaker verification (ASV), and speaker diarization (SD) \citep{chang2022distilhubert}.

To structure the literature review, the following sections are presented as follows. Section \ref{sec:self_supervised} presents an overview of recent models for self-supervised representation learning. Then, efficiency-related approaches are organized into the optimization of existing models (Section \ref{sec:optimization_existing}), architecture efficiency (Section \ref{sec:architecture_efficiency}), fine-tuning efficiency (Section \ref{sec:finetuning_efficiency}), and data efficiency (Section \ref{sec:data_efficiency}). Finally, section \ref{sec:conclusions} presents conclusions and future directions.

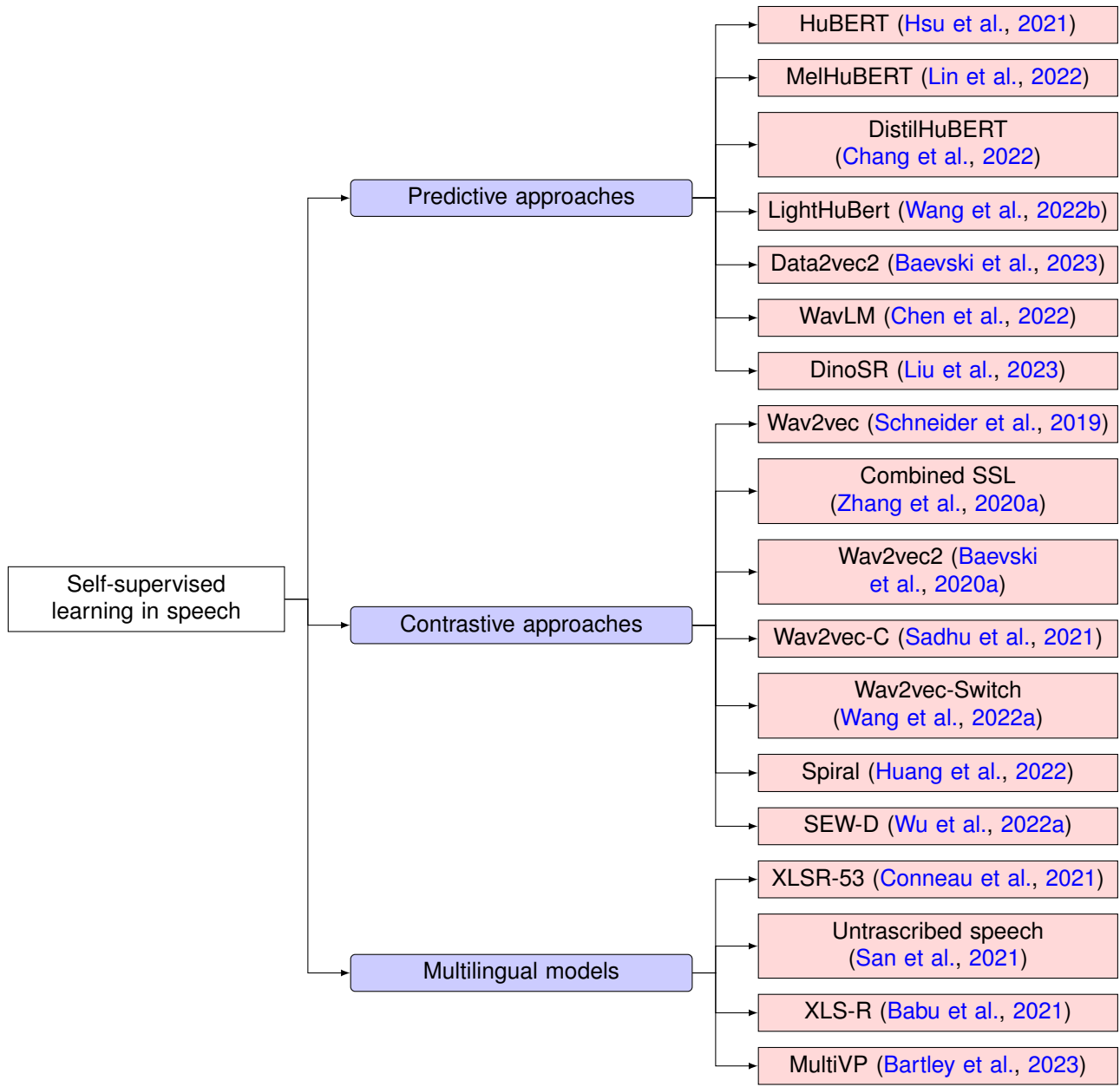
\begin{figure*}
\centering
\tikzset{
  basic/.style  = {draw, text width=4cm, align=center, font=\sffamily, rectangle},
  root/.style   = {basic, rounded corners=2pt, thin, align=center, fill=green!30},
  onode/.style = {basic, thin, rounded corners=2pt, align=center, fill=green!60,text width=3cm,},
  tnode/.style = {basic, thin, align=left, fill=pink!60, text width=15em, align=center},
  xnode/.style = {basic, thin, rounded corners=2pt, align=center, fill=blue!20,text width=5cm,},
  wnode/.style = {basic, thin, align=left, fill=pink!10!blue!80!red!10, text width=6.5em},
  edge from parent/.style={draw=black, edge from parent fork right}
}
\begin{forest} for tree={
  grow=east,
  growth parent anchor=west,
  parent anchor=east,
  child anchor=west,
  edge path={\noexpand\path[\forestoption{edge},->, >={latex}] 
        (!u.parent anchor) -- +(10pt,0pt) |-  (.child anchor) 
        \forestoption{edge label};}
}
[Self-supervised learning in speech, basic,  l sep=10mm,
  [Multilingual models, xnode,  l sep=10mm,
      [MultiVP \citep{bartley2023accidental}, tnode]
      [XLS-R \citep{babu2021xls}, tnode]
      [Untrascribed speech \citep{san2021leveraging}, tnode]
      [XLSR-53 \citep{conneau2021unsupervised}, tnode] ]
  [Contrastive approaches, xnode,  l sep=10mm,
      [SEW-D \citep{wu2022performance}, tnode]
      [Spiral \citep{huang2022spiral}, tnode]
      [Wav2vec-Switch \citep{wang2022wav2vec}, tnode]
      [Wav2vec-C \citep{sadhu2021wav2vec}, tnode]
      [Wav2vec2 \citep{baevski2020wav2vec}, tnode]
      [Combined SSL \citep{zhang2020pushing}, tnode]
      [Wav2vec \citep{schneider2019wav2vec}, tnode] ]
  [Predictive approaches, xnode,  l sep=10mm,
      [DinoSR \citep{liu2023dinosr}, tnode]
      [WavLM \citep{chen2022wavlm}, tnode]
      [Data2vec2 \citep{baevski2023efficient}, tnode]
      [LightHuBert \citep{wang2022lighthubert}, tnode]
      [DistilHuBERT \citep{chang2022distilhubert}, tnode]
      [MelHuBERT \citep{lin2022melhubert}, tnode]
      [HuBERT \citep{hsu2021hubert}, tnode] ] ]
\end{forest}
\caption{Recent self-supervised approaches for speech representation learning, including contrastive and predictive approaches, and multilingual models.}
\label{fig:self_supervised_tree}
\end{figure*}

\section{Self-supervised learning for speech} \label{sec:self_supervised}

Most self-supervised learning architectures for speech processing use a combination of convolutional and self-attention layers. Overall, convolutional layers are good at getting local features. They are widely used in vision because they can model edges and shapes by extracting positional information from local features. For global modeling, however, convolutional layers are limited. That explains why convolutional networks need a large stack of layers to grasp global features. Transformers, on the other hand, are good at modeling global context. They are also easier to train in parallel \citep{gulati2020conformer}.

Recent work has been done with transformer architectures to deal with local feature limitations by adding positional information \citep{wu2020lite,yang2019convolutional,yu2018qanet}. Another alternative is to use a transformer channel and a convolutional channel, concatenating at the end the output of the two channels \citep{bello2019attention}. For speech processing, a neural configuration known as a Convolution-augmented Transformer, or Conformer \citep{gulati2020conformer}, organically combines convolutional layers and self-attention layers from transformers, using feed-forward layers in a macaron configuration to wrap intermediate layers. Also, multiple heads for attention enable the model to focus on different parts of the input sequences.

In particular, two approaches have emerged from recent self-supervised speech representation models: Hidden unit BERT (HuBERT) \citep{hsu2021hubert} and wav2vec2 \citep{baevski2020wav2vec}. The latter emerged because of its solid performance in speech representation and downstream tasks, paired with the availability of pretrained multilingual models. The former emerged because of its simplicity and stability, as well as the way it trains the neural model, which is similar to classical frame level ASR, spearheading further work from research extensions  \citep{mohamed2022self}.

Both HuBERT and wav2vec2 combine convolutional and self-attention layers in a dual channel configuration, where a teacher channel generates latent representations, and a student channel learns to replicate the latent space during pretraining. For the pretraining loss, HuBERT uses a predictive loss, while wav2vec2 uses a contrastive loss. Therefore, based on the pretraining loss, we can group self-supervised speech representation approaches into predictive and contrastive methods (Figure \ref{fig:self_supervised_tree}).

\subsection{Predictive self-supervised approaches}

Self-supervised models for speech representations face three challenges. First, input utterances have variable numbers of sound inputs. Secondly, there is no lexicon for input sounds. Thirdly, input sounds are not clearly segmented in the input utterances, as speech data is variable \citep{hsu2021hubert}. Several models have been recently proposed to address these challenges (Figure \ref{fig:self_supervised_tree}), using a predictive loss during pretraining.

HuBERT \citep{hsu2021hubert} uses a predictive loss for self-supervised representation learning. It also uses masked segments of input data for pretraining, similar to the training of language models with masked words in large natural language datasets \citep{devlin2019bert}. Formally, the loss is calculated as follows \citep{hsu2021hubert}:

\begin{equation}
  X = \left[x_1, \cdots x_T \right] 
\end{equation}
\begin{equation}
  \tilde{X} = r(X, M)
\end{equation}
\begin{equation}
  z_t = h(x_t)
\end{equation}
\begin{equation}
  \mathcal{L} = \sum_{t \in M} \log p_f(z_t | \tilde{X}, t) 
\end{equation}

where $X$ is an input utterance of length $T$, $M$ is the set of indices to be masked, $r$ replaces $x_t$ with a mask embedding $\tilde{x}$ if $t \in M$, $h$ is a clustering model, and $f$ is a masked prediction model that predicts a distribution $p_f$ over the target indices at each timestep.   

By learning randomly masked sounds from the input utterances, HuBERT learns both the speech representation and the language model, which is closely related to the long-term relationships between inputs, as such long-term relationships come from the language itself. The loss is calculated only on the masked inputs so that the model learns the two features at the same time \citep{hsu2021hubert}. 

The neural architecture of HuBERT comprises a convolutional module, a self-attention module, and a projection module. For the self-supervised training, labels come from the cluster assignments that kmeans generates. The model should learn to match the cluster assignments from kmeans during pretraining. Overall, the training involves three stages. The first stage trains the model with kmeans assignments from Mel-frequency cepstral coefficients (MFCC). The second stage trains the model with kmeans assignments from the representation learned in the first stage. The third stage performs fine-tuning with ASR as the downstream speech application, replacing the projection module with a softmax layer for the connectionist temporal classification (CTC) loss, and freezing the convolutional part of the architecture. Metrics for the self-supervised training iterations include phoneme purity, entropy, and phoneme normalized mutual information. During the supervised fine-tuning, HuBERT uses the word error rate \citep{hsu2021hubert}.

WavLM \citep{chen2022wavlm} is an approach very similar to HuBERT. It emphasizes spoken content modeling and the preservation of speaker identity. Most methods use utterances containing speech from a single speaker for pretraining. In contrast, WavLM combines utterances from different speakers to perform data augmentation of the input sequence. it also extends the self-attention mechanism with gated relative position bias to improve the quality of the learned representations \citep{chen2022wavlm,mohamed2022self}.

DinoSR (Self-distillation and online clustering for self-supervised representation learning) replaces the kmeans in HuBERT with an online clustering approach \citep{liu2023dinosr}. It combines masked input pretraining and online clustering for target discretization. Similar to other self-training approaches, the teacher weights are updated using the exponential moving average, while the student uses backpropagation to adjust the weights during training. There is no need for approximation in targets -- like wav2vec2 or similar methods -- because the online clustering targets for the teacher do not need to be backpropagated. DinoSR results show improvements in ASR against base models of previous methods \citep{baevski2020wav2vec,hsu2021hubert}, as well as improvements in SUPERB \citep{ yang2021superb} results.

Data2vec2 \citep{baevski2023efficient} is a teacher -- student model trained with a predictive loss. The teacher weights are updated from the student weights using an exponential moving average for every update. The proposed model supports a multimodal input, using the same loss for image, text, or speech. Feature extractors change to adapt the model to the input data, but the core transformer module remains the same.

To improve the training efficiency of Data2vec2, the training relies on contextualized training, using the whole sample to provide context to the model. Training also reuses the target representation from the teacher for every masked version of the student input. The neural architecture has a fast convolutional decoder and efficient data encoding. The predictive loss is calculated from the L2 distance between the student output for the masked input and the target output the teacher generates for the unmasked input sample \citep{baevski2023efficient}.

Data2vec2 can be up to ten times faster than existing self-learning representation models in speech, but it still manages to have a better performance. The base model is pre-trained using LibriSpeech \citep{panayotov2015librispeech}, while the large one uses LibriLight \citep{kahn2020libri}, fine-tuning for ASR and using a 4-gram language model during decoding.

\subsection{Contrastive self-supervised approaches}

Contrastive approaches replace the predictive loss during self-training for a contrastive loss. A contrastive loss in a teacher -- student configuration is defined as follows \citep{chen2020simple,huang2022spiral}:

\begin{equation}
  \phi (a, b) = \frac{a^T b}{\left\lVert a \right\rVert \left\lVert b \right\rVert} 
\end{equation}
\begin{equation}
  \mathcal{L} = - \sum_{i = 1}^{T} \log \frac{e^{\phi(z_i, z'_i)/\tau}}{\sum_{j \in D_i}e^{\phi(z_i, z'_j)/\tau}} 
\end{equation}

where $z$ is the latent representation from the student network and $z'$ is the latent representation from the teacher network, $\tau$ represents a temperature parameter, and $D_i$ is the set of distractors for the $z_i$ representation. 

Wav2vec2 \citep{baevski2020wav2vec} is a popular approach in the contrastive loss category. It combines masked input timesteps and quantization through Gumbel SoftMax \citep{jangcategorical} to learn latent speech representations. Follow-up studies of wav2vec2 include wav2vec-C \citep{sadhu2021wav2vec}, which extends the loss of wav2vec2 with a consistency term to reconstruct the input from quantized representations, and wav2vec-switch \citep{wang2022wav2vec}, which adds noise robustness to the contextualized representation learning of wav2vec2.

Self-supervised perturbation invariant representation learning (SPIRAL) also pretrains the model using a contrastive loss \citep{huang2022spiral}. SPIRAL comprises a subsampling convolutional module, a self-attention module, a second subsampling convolutional module, and a second self-attention module. Then, a projection module generates the architecture output. An additional predictor module for the student only improves performance. Both teacher and student have the same architecture, except for the student predictor module. SPIRAL gets results similar to or better than wav2vec2. It, however, uses no quantization and considerably reduces training costs.

The input to the student part of the model is perturbed with noise, and the student's objective is to learn a denoising representation as close as possible to the teacher output, considering that the teacher input has no noise added to it. The utterance representation uses log Mel filter banks with 128 filters. In addition to SpecAugment and dropout \citep{park2019specaugment,park2020specaugment}, the input is also regularized by noise addition during pretraining, using a method known as multi-condition pretraining \citep{chiba2019multi,seltzer2013investigation}, where noise from a dataset is added to the input utterance avec varying degrees of signal to noise ratio. To avoid learning trivial weights, SPIRAL uses two methods. First, it randomly adds padding to the teacher's input, avoiding any positional information becoming the student's learning objective. Secondly, it uses an in-utterance contrastive loss, mixing the negative samples in the learning objective \citep{huang2022spiral}.

Combined SSL improves the state-of-the-art in ASR by combining a conformer architecture, wav2vec2 pretraining, and noisy student self-training, where SpecAugment provides the noise for the input speech sequences \citep{zhang2020pushing}. Combined SSL uses a convolutional subsampling module, a linear module, and a conformer block. The convolutional subsampling module can be seen as a feature encoder, while the conformer module serves as a context network. The positional embeddings of the self-attention are removed with no negative impact on the final results. Removing positional embeddings has no negative impact because the convolutional module learns relative positional information, according to previous research \citep{wang2020transformer}.

For Combined SSL pretraining, the original raw waveforms in wav2vec2 are replaced by 80 Mel log spectral coefficients, which makes the input sequence shorter. Another variation against wav2vec2 is the removal of the quantization module, using a linear layer instead. CombinedSSL uses 512 TPUs for the largest model during four days. LibriSpeech is the dataset to test supervised ASR, while Librilight provides unlabeled data for petraining.

\begin{table*}[t]
  \centering
  \setlength{\extrarowheight}{.5em} 
  \begin{tabu} to 0.9\textwidth { X[4,l] X[1.5,c] X[l] X[c] X[c] }
  \hline
    \textbf{Model} &
    \textbf{WER (\%)} &
    \textbf{GPUs} &
    \textbf{Size (M)} &
    \textbf{Hours} \\\hline
  \hline
    Transformer Acoustic \citep{wang2020transformer} &
    4.8 &
    32 &
    93 &
    96 \\
  
    BLSTM Acoustic \citep{wang2020transformer} &
    6.6 &
    32 &
    163 &
    96 \\

    Hubert Large \citep{hsu2021hubert} &
    3.3 &
    128 &
    300 &
    99.75 \\

    Hubert XLarge \citep{hsu2021hubert} &
    2.9 &
    256 &
    1000 &
    99.75 \\
    
    wav2vec2 Base \citep{baevski2020wav2vec} &
    4.8 &
    64 &
    95 &
    38.4 \\

    wav2vec2 Large \citep{baevski2020wav2vec} &
    3.3 &
    128 &
    317 &
    180 \\

    Combined SSL XXL \citep{zhang2020pushing} &
    2.7 &
    512 $^{*}$ &
    1000 &
    168 \\

    Combined SSL XXL+ \citep{zhang2020pushing} &
    2.6 &
    512 $^{*}$ &
    1050 &
    168 \\

    Data2vec2 Base \citep{baevski2022data2vec} &
    5.2 &
    16 &
    -- &
    43.3 \\

    Data2vec2 Large \citep{baevski2023efficient} &
    3.5 &
    64 &
    -- &
    76.7 \\

    Spiral Base \citep{huang2022spiral} &
    6.1 &
    16 &
    95 &
    31.2 \\

    Spiral Large \citep{huang2022spiral} &
    3.5 &
    32 &
    317 &
    174 \\

    DinoSR \citep{liu2023dinosr} &
    6.7 &
    16 &
    -- &
    180 \\

    \hline
  \end{tabu}
  \caption{Recent models for self-supervised speech representation learning. Results are for the LibriSpeech test-other dataset, using WER as the evaluation metric, and ASR as the downstream task for fine tuning the neural architectures. $(^{*})$ This model uses TPUs.}
  \label{table:self_supervised_tradeoff}
  \end{table*}

\subsection{Multilingual models}

Multilingual approaches are particularly important for languages and dialects without written resources for training, as cross-lingual transfer facilitates learning speech representations in low-resource settings \citep{abdullah2023nature,san2021leveraging}. Additionally, cross-lingual pretraining outperforms monolingual pretraining after fine-tuning for speech processing tasks, particularly in low resource languages \citep{conneau2021unsupervised}.

The wav2vec2 architecture serves as the base for several multilingual approaches, including XLS-R \citep{babu2021xls} and XLSR-53 \citep{conneau2021unsupervised}. XLSR-53 pretrains a large architecture with unlabeled speech data in multiple languages. It uses a shared quantization module, which forces the model to share discrete tokens across languages, learning bridges between different languages. The model supports 53 languages. XLS-R extends the training of XLSR-53 going from 56 thousand hours to nearly half a million hours of publicly available speech data. Language coverage also improves, going from 53 languages to 128 languages, and evaluation is performed with various speech processing tasks, including language identification, ASR, and speech translation. 

MultiVP \citep{bartley2023accidental} pretrains a conformer-based model with a multilingual dataset, fine-tuning it to perform spoken language identification. Lower layers of the model encode enough information to discriminate languages and identify speech sounds. The fine-tuned model gets similar results to XLSR in language identification, though it uses only around 20\% of parameters. MultiVP can also identify unseen languages, showing its ability to realize zero-shot language identification.

\subsection{Limitations of self-supervised models}

The word error rate (WER) is widely used in evaluating speech representation models after supervised fine-tuning for ASR (Table \ref{table:self_supervised_tradeoff}). The error rate is the relationship between correct input-output words and the testing set size \citep{graves2012supervised}:

\begin{equation}\label{eq:wer}
WER =  \frac{1}{| S' |} \sum_{(x,z) \in S'} 
  \left\{ 
  \begin{array}{l l}
  0 \text{  if } h(x)=z \\
  1 \text{  otherwise }
  \end{array} 
  \right.
\end{equation}

where $S'$ is the test set, $x$ is the input, $z$ is the expected output, and $h$ is the learning model.

But alternative metrics have been proposed to improve the evaluation of speech representations, including $ABX$ and $D_{SSIMI}$ \citep{chung2018speech2vec,dunbar2021zero,kahn2020libri,nguyen2020zero}. $D_{SSIMI}$ calculates the distance between a pair of words from a semantic perspective, correlating the distance of learned representations with a human judgment of semantic similarity. On the other hand, $ABX$ is an acoustic -- phonetic metric. It measures how well separated phonetic categories are in a latent representation space \citep{nguyen2020zero}. For a given pair of sounds, we calculate the probability that two sounds of the same category are closer to one another than two sounds of different categories. For a triphone $(a, b, x)$, from categories $A$ and $B$, we want to know the probability of

\begin{equation}
  d(a, x) < d(b, x)
\end{equation}
\begin{equation}
  a \in A, b \in B, x \neq a \in A
\end{equation}

where $d$ is a framewise distance. It could be the Kullback-Leibler or the angular distance \citep{nguyen2020zero}. 

Though the evaluation of speech representation models is mostly done with ASR, some datasets have been proposed to test representation models with other speech processing tasks. Those datasets include speech processing universal performance benchmark (SUPERB) and LeBenchmark \citep{evain2021lebenchmark,yang2021superb}. SUPERB includes phoneme recognition, keyword spotting, intent classification, speaker identification, emotion recognition, automatic speech recognition, query by example spoken term detection, slot filling, automatic speaker verification, and speaker diarization \citep{chang2022distilhubert}.

Additionally, few existing models use half-precision floating point numbers, even though this technique can half the memory requirements and accelerate the arithmetic computations on recent GPUs \citep{micikevicius2018mixed}. A similar issue happens with dynamic batching \citep{tyagi2020taming}, a procedure that avoids wasting computing resources on the padded portion of speech mini-batches.

\begin{figure}[htb]
\begin{minipage}[b]{1.0\linewidth}
  \centering
  \centerline{\includegraphics[width=9.5cm]{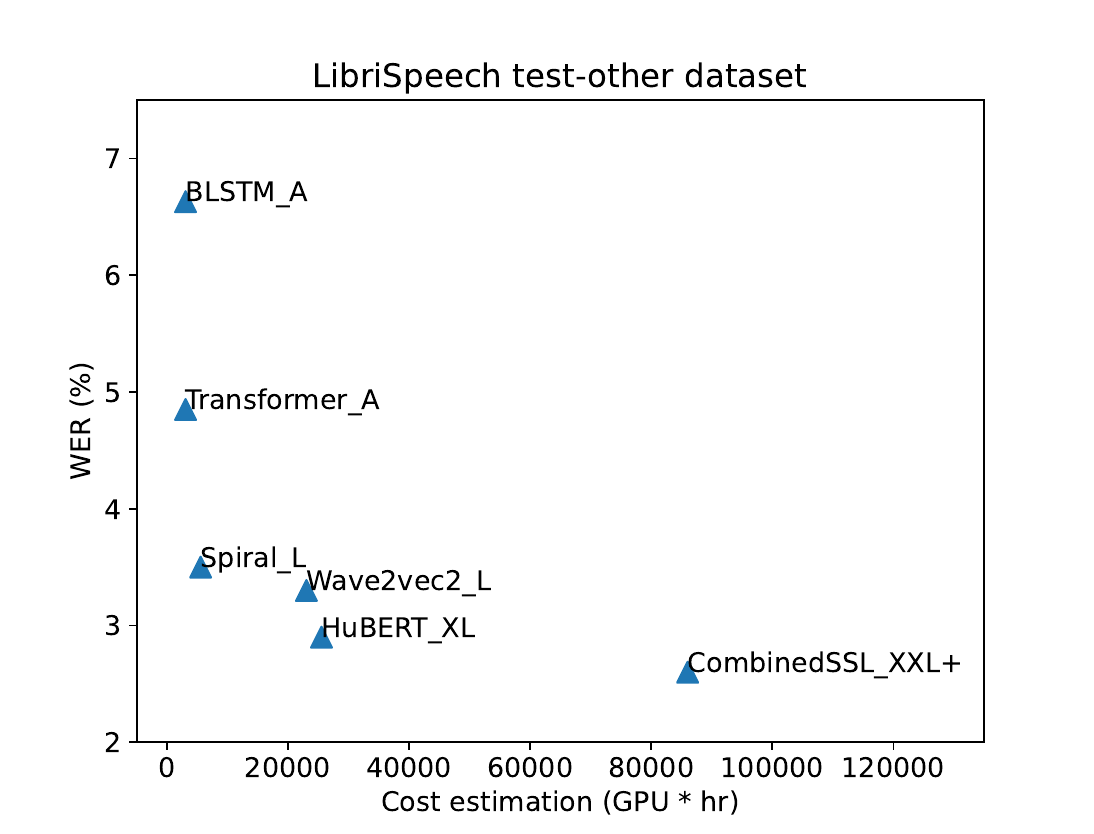}}
 \vspace{1.0cm}
\end{minipage}
\caption{Existing models for self-supervised speech representation learning. This trend illustrates the trade-off between computational costs and representation performance.}
\label{fig:librispeec_test_other_wer}
\end{figure}

Despite the groundbreaking results in speech processing, self-supervised models are computationally expensive. Large training costs represent a challenge. Indeed, there is a trade-off between self-training computational costs and speech representation performance (Figure \ref{fig:librispeec_test_other_wer}). For example, the best-performing method, Combined SSL \citep{zhang2020pushing}, uses 512 Tensor Processing Units (TPUs) for four days to train its XXL+ version. In contrast, BLSTM \citep{wang2020transformer} requires only 32 GPUs for 96 hours, but its representation performance degrades, passing from a WER of 2.6\% to 6.63\% for the LibriSpeech test-other dataset (Table \ref{table:self_supervised_tradeoff}). 

The computational cost of large self-supervised models hinders the study of other training recipes and the reproduction of experimental results, as few researchers can afford the cost \citep{lin2022melhubert}. Computational costs limit the number of laboratories with enough computational resources to perform research, which creates a barrier to developing new approaches \citep{parcollet2023efficiency,wang2023read}. Computational costs also have environmental implications, as training requires considerable amounts of energy \citep{parcollet2023efficiency,wu2022sustainable}.

Current efforts to deal with computational costs in large self-supervised models include fine-tuning with parameter-efficient transfer learning (PETL) \citep{hu2021lora,zhang2023adaptive}, optimization of neural architecture components \citep{dao2022flashattention,lee2022fnet,moumen2023stabilising,wu2020lite}, and optimization of existing speech models \citep{chang2022distilhubert,lin2022melhubert}.

\section{Towards optimization of existing speech models}\label{sec:optimization_existing}

As mentioned, self-supervised models for speech representation have generated impressive results in multiple downstream speech processing tasks. However, those models are quite large, requiring a lot of computational resources for training. One alternative to reduce model size is knowledge distillation \citep{allen2020towards}, where a small student model learns from a large teacher model, which has been pretrained and remains frozen during student training \citep{peng2023dphubert}.

Drawing inspiration from DistilBERT \citep{sanh2019distilbert}, DistilHuBERT \citep{chang2022distilhubert} is a knowledge distillation approach that reduces the model size by 75\% and speeds up inference by 73\%. DistilHuBERT comprises a convolutional module with seven layers, two self-attention layers, and three multitask layers, which learn to predict layers 4, 8, and 12 from the transformer in HuBERT. These layers were chosen because close layers tend to encode similar information. Layers in DistilHuBERT are initialized by their counterparts in the teacher module. Training requires a single GPU for 200k iterations, which takes roughly 55 hours, with a model size of 24M parameters, compared to the 95M from HuBERT. For the loss, DistilHuBERT optimizes both the L1 distance and the cosine similarity, using a lambda parameter as a factor of the cosine similarity.

For the speech processing tasks, a weighted sum of the multitask layers generates the output of the DistilHuBERT architecture. Results are comparable to those of the teacher model, demonstrating the ability of the small student model to capture the HuBERT speech representation while pretraining with the LibriSpeech dataset and testing with SUPERB \citep{chang2022distilhubert}.

LightHuBERT \citep{wang2022lighthubert} also aims at optimizing the HuBERT architecture. It relies on knowledge distillation to learn a once-for-all transformer model. The teacher is a HuBERT base model, while the student learns by predicting masked inputs in an iterative process. The transformer in LightHuBERT comprises subnets with sharable weights and several configuration parameters, enabling a random search to adjust the model to different resource constraints. Results outperform the teacher model and improve DistilHuBERT performance, while the number of parameters is reduced by up to 29\% against the base model.

Similarly, FitHuBERT \citep{lee2022fithubert} proposes a thinner and deeper neural configuration than HuBERT, without reducing the number of layers. Training is performed with knowledge distillation from a pretrained HuBERT model, reducing the model size and inference time.

The student architecture in knowledge distillation methods is manually designed. And it does not change during training. However, modifying student architectures can have a considerable impact on model results, even for student architectures with similar sizes \citep{ashihara2022deep}. Therefore, a joined distillation and pruning approach for speech SSL has been recently proposed, using HuBERT (DPHuBERT) or WavLM (DPWavLM) as the teacher models \citep{peng2023dphubert}. The pruning performed during training is structured, which enables the removal of groups of parameters -- an entire layer, for instance -- from the base model \citep{peng2023structured}. In contrast, unstructured pruning sets to zero individual weights in the model, generating a sparse architecture without speedup. The joined approach improves over previous knowledge distillation approaches in most tasks of the SUPERB dataset \citep{peng2023dphubert}.    

Another approach to reduce the cost of self-supervised learning is MelHuBERT \citep{lin2022melhubert}. Contrary to DistilHuBERT, DPHuBERT, or DPWavLM, MelHuBERT does not use knowledge distillation. Instead, it is a simplified version of HuBERT that has twelve self-attention layers and a weighted sum of all the layers for downstream tasks, as each self-attention layer best encodes a certain aspect of the input speech sequence \citep{pasad2021layer,pasad2023comparative}. For example, the sixth layer best represents phoneme information and cluster purity. Though results degrade from the original HuBERT model in terms of phoneme recognition and speaker identification, they remain better than other existing models such as wav2vec \citep{schneider2019wav2vec}, vq-wav2vec \citep{baevski2020vqwav2vec}, and multilayer LSTMs \citep{yeh2022autoregressive}.

MelHuBERT reduces the multiplication and addition calculations (MAC) by 33\%. As the input is a 40-dimensional Mel log spectrogram, input sequences are shorter. Additionally, the spectrogram calculation removes the need for the convolutional module for feature extraction, further reducing the model size. MelHuBERT also reduces the iterative training scheme, relying only on two training stages. Training of the model requires a single GPU during 150hrs, using a subset of LibriSpeech \citep{lin2022melhubert}.

It is also possible to train HuBERT in an academic setup \citep{chen2023reducing}, using only 8 GPUs to train the model. This approach gets similar results to the original model. Optimizations include mixed precision training and gradient accumulation to match the batch size of the original model, keeping in mind that the batch size has a considerable impact on model results. 

Along with optimizations to modify the HuBERT architecture, there are also efforts to optimize the wav2vec architecture. Proposed approaches optimizing wav2vec include squeezed and efficient wav2vec2 with disentangled attention (SEW-D) \citep{wu2022performance}, stochastic squeezed and efficient wav2vec2 (S-SEW) \citep{vyas2022demand}, and PARP (Prune Adjust and Re-Prune) \citep{lai2021parp}.

PARP \citep{lai2021parp} prunes a pretrained wav2vec2 model to improve its computational complexity, obtaining a sparsity mask and updating it while finetuning the speech representation for downstream tasks.

In contrast, SEW-D \citep{wu2022performance} proposes four changes for optimizing wave2vec2, reducing inference and pretraining time: a squeezed context module, a compact feature extractor, MLP projection heads, and a disentangled attention instead of multiheaded attention. Disentangled attention uses relative positional embeddings and context embeddings separately, combining them in the weight calculation with content to content, content to positional, and positional to content attention. This disentangled configuration outperforms multiheaded attention in transformer models.   

Instead of a linear projection layer, SEW-D uses a projection MLP with two linear layers, each one with ReLU activations and batch normalization. The compact feature extractor decreases the number of convolutional filters in the wav2vec2 original architecture, reducing the size of the first convolutional layer and increasing subsequent layers' size as the frame rate decreases. This convolutional configuration helps to more evenly distribute the computations in the forward and backward pass. After the compact feature extractor, the squeezed context module takes as input a downsampled tensor, processes it, and upsamples the output before calculating the pretraining loss. This squeeze reduces the computations in the transformer module, improving overall performance \citep{wu2022performance}. 

The base configurations of SEW-D are pretrained for 100k iterations and finetuned for 80k iterations. Pretraining takes a day with 8 GPUs. For comparison against wav2vec2 ASR results, pretraining goes up to 400k iterations. Optimizations allow SEW-D to get between 2 to 3 times faster inference against wav2vec2 baseline configurations. It also has faster pretraining times, using 8 GPUs for 24 hours in base configurations and getting comparable or better word error rates \citep{wu2022performance}.

Though SEW-D reduces wave2vec2 complexity, the performance degrades. Thus, S-SEW \citep{vyas2022demand} proposes a probabilistic approach to smooth the trade-off between performance and computational complexity. To do so, the squeezing factor in SEW is replaced with a stochastic factor, choosing it randomly for each iteration from a uniform distribution. S-SEW also performs a stochastic pooling of key and value in the self-attention layers, which further improves computational efficiency.

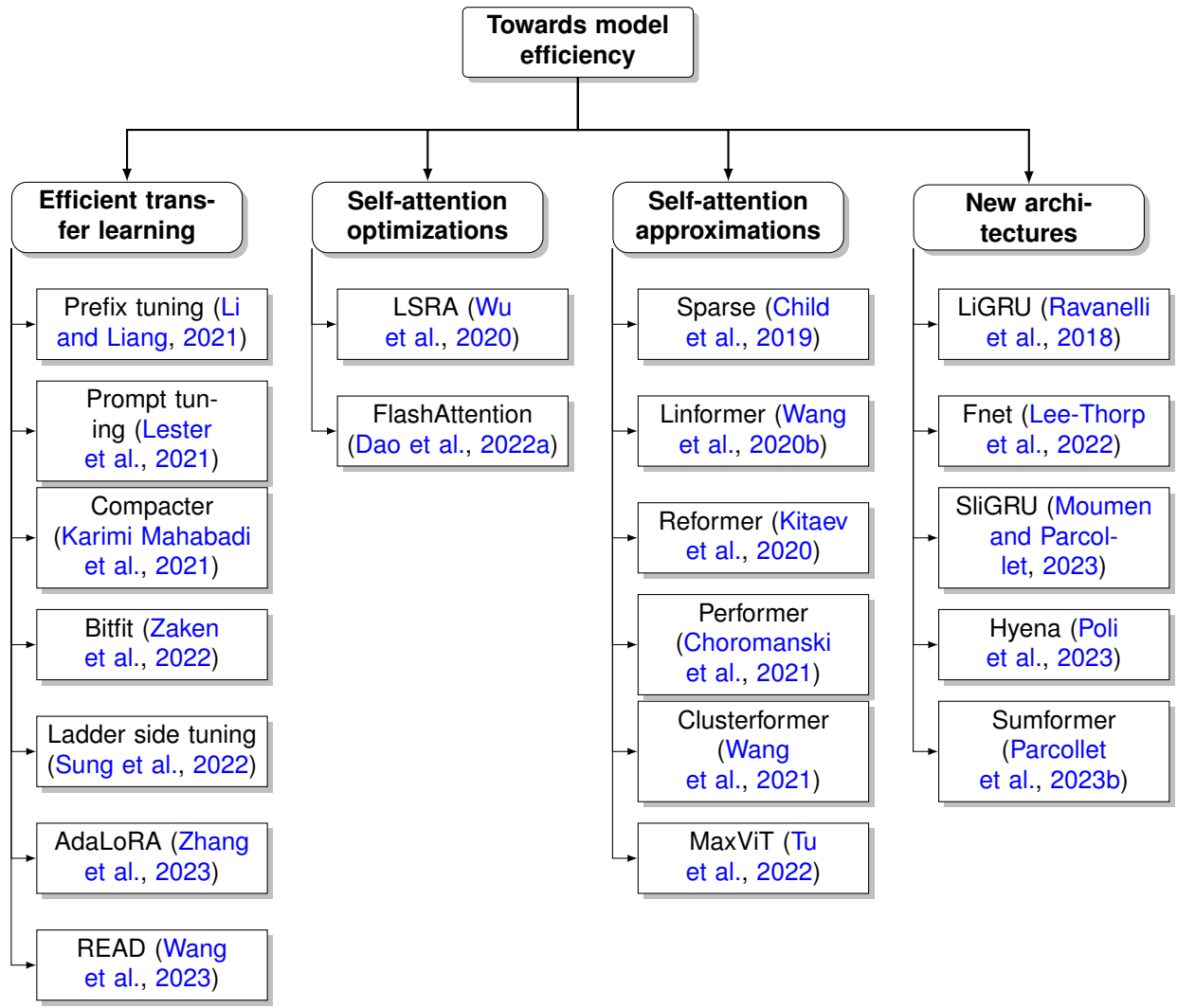
\begin{figure*}
\centering

\tikzset{
  basic/.style  = {draw, text width=3cm, drop shadow, font=\sffamily, rectangle},
  root/.style   = {basic, rounded corners=2pt, thin, align=center, fill=white},
  level-2/.style = {basic, rounded corners=6pt, thin,align=center, fill=white, text width=3cm},
  level-3/.style = {basic, thin, align=center, fill=white, text width=3.0cm}
}

\begin{tikzpicture}[
  level 1/.style={sibling distance=12em, level distance=7em},
  edge from parent/.style={->,solid,black,thick,sloped,draw}, 
  edge from parent path={(\tikzparentnode.south) -- (\tikzchildnode.north)},
  >=latex, node distance=1.5cm, edge from parent fork down]

\node[root] {\textbf{Towards model efficiency}}
  child {node[level-2] (c1) {\textbf{Efficient transfer learning}}}
  child {node[level-2] (c2) {\textbf{Self-attention optimizations}}}
  child {node[level-2] (c3) {\textbf{Self-attention approximations}}}
  child {node[level-2] (c4) {\textbf{New architectures}}};

\begin{scope}[every node/.style={level-3}]
\node [below of = c1, xshift=10pt] (c11) {Prefix tuning \citep{li2021prefix}};
\node [below of = c11] (c12) {Prompt tuning \citep{lester2021power}};
\node [below of = c12] (c13) {Compacter \citep{karimi2021compacter}};
\node [below of = c13] (c14) {Bitfit \citep{zaken2022bitfit}};
\node [below of = c14] (c15) {Ladder side tuning \citep{sung2022lst}};
\node [below of = c15] (c16) {AdaLoRA \citep{zhang2023adaptive}};
\node [below of = c16] (c17) {READ \citep{wang2023read}};

\node [below of = c2, xshift=10pt] (c21) {LSRA \citep{wu2020lite}};
\node [below of = c21] (c22) {FlashAttention \citep{dao2022flashattention}};

\node [below of = c3, xshift=10pt] (c31) {Sparse \citep{child2019generating}};
\node [below of = c31] (c32) {Linformer \citep{wang2020linformer}};
\node [below of = c32] (c33) {Reformer \citep{kitaev2020reformer}};
\node [below of = c33] (c34) {Performer \citep{choromanski2021rethinking}};
\node [below of = c34] (c35) {Clusterformer \citep{wang2021cluster}};
\node [below of = c35] (c36) {MaxViT \citep{tu2022maxvit}};

\node [below of = c4, xshift=10pt] (c41) {LiGRU \citep{ravanelli2018light}};
\node [below of = c41] (c42) {Fnet \citep{lee2022fnet}};
\node [below of = c42] (c43) {SliGRU \citep{moumen2023stabilising}};
\node [below of = c43] (c44) {Hyena \citep{poli2023hyena}};
\node [below of = c44] (c45) {Sumformer \citep{parcollet2023sumformer}};
\end{scope}

\foreach \value in {1,...,7}
  \draw[->] (c1.195) |- (c1\value.west);

\foreach \value in {1,2}
  \draw[->] (c2.195) |- (c2\value.west);

\foreach \value in {1,...,6}
  \draw[->] (c3.195) |- (c3\value.west);
  
\foreach \value in {1,...,5}
  \draw[->] (c4.195) |- (c4\value.west);
\end{tikzpicture}

\caption{Methods proposing efficiency-oriented approaches that can deal with the computational costs of self-supervised learning architectures, including new architectures, self-attention improvements, and efficient transfer learning.}
\label{fig:towards_efficiency}
\end{figure*}
  
\section{Towards neural architecture efficiency} \label{sec:architecture_efficiency}

In general, self-attention layers have contributed to breakthrough results in modeling speech, language, vision, biology, and other domains \citep{mohamed2022self,poli2023hyena}. Self-attention layers can generalize to unseen data and tasks because of their capacity for in-context learning. They can also learn dependencies from the whole input without any restrictions in context. Self-attention scales quadratically to sequence length, becoming increasingly expensive as the input length grows. Recent work shows they use only a small portion of their quadratic capabilities when processing language \citep{poli2023hyena}.

Sub-quadratic alternatives for self-attention layers in transformers have been proposed in recent work (Figure \ref{fig:towards_efficiency}). These sub quadratic layers use linear \citep{schlag2021linear,zhai2021attention}, learnable \citep{kitaev2020reformer,wang2021cluster}, sparse \citep{child2019generating,roy2021efficient,tu2022maxvit}, or low-rank \citep{choromanski2021rethinking,wang2020linformer} approximations to accelerate layer calculations \citep{poli2023hyena,tay2022efficient}. Though these approximations improve the efficiency of self-attention layers by helping alleviate their quadratic relationship to the input sequence length, they reduce the quality of the models \citep{dao2022flashattention}, requiring a hybrid combination with standard self-attention layers to get a similar quality \citep{dao2022hungry,mehta2022long,poli2023hyena}. 

But some architectures scale subquadratically to sequence length without compromising model quality. They include the Sumformer \citep{ parcollet2023sumformer}, the Fnet \citep{lee2022fnet}, and the Hyena hierarchy \citep{poli2023hyena}.

The Hyena hierarchy \citep{poli2023hyena} is an architecture that implements a recurrence of two subquadratic primitives: a long convolution and an element-wise multiplication. The number of recurrences sets the size of the operator. Results match transformer models, but computational costs are lower and the model is faster -- especially at long sequences. This improvement happens because of its sub-quadratic scaling in terms of sequence length.

The Fnet \citep{lee2022fnet} replaces self-attention completely. Instead, it uses the Fast Fourier transform (FFT) plus a feed-forward layer. This change speeds up the computations because FFTs are fast to calculate in GPUs. It also reduces the number of network parameters because the learnable matrices in self-attention layers are no longer necessary. The efficiency boost from this change can go up to 97\% against models using self-attention layers, while the degradation in modeling results remains small.

Similar to Fnets, Sumformers \citep{ parcollet2023sumformer} replace the self-attention layers altogether. Weights in self-attention layers of some speech processing architectures behave close to a mean function, suggesting self-attention is not essential for speech recognition architectures processing acoustic utterances. Instead of using multiheaded self-attention, it is possible to replace it with a linear alternative, which calculates the mean vector of all time steps in a speech utterance. Time-specific information complements the mean vector, generating an approach known as summary mixing. Using summary mixing instead of multiheaded self-attention, the sumformer processes speech data faster than existing speech processing approaches without degradation in modeling performance. The sumformer generates up to 27\% improvement in training and inference time. It also reduces in half the RAM requirements.

Other methods optimize self-attention layers with long short range attention in a lite transformer \citep{wu2020lite}, or GPU efficient operations in FlashAttention \citep{dao2022flashattention}, without using approximations. 

FlashAttention \citep{dao2022flashattention} improves self-attention layers' efficiency by optimizing the input -- output (IO) memory operations in the GPU. Overall, GPUs have two kinds of memories. A small SRAM is associated with each kernel, while a large, slower, high-bandwidth memory is shared between all the kernels. Memory-intensive operations, like the matrix operation of the self-attention layers, have their bottleneck at the read-write RAM access. In contrast, compute-intensive operations have their bottleneck in the number of arithmetic operations that must be realized. As self-attention is primarily a memory-intensive operation, FlashAttention reduces the number of IO operations by tiling, assigning a matrix operation to a single kernel, and saving some results from the forward pass to share in the subsequent backward pass. 

The optimization of IO operations can speed up models using FlashAttention by three times. The impact of the optimization is higher in long sequences. It is also possible to insert input approximations, such as sparsity, to combine memory optimization and input length reduction\citep{dao2022flashattention}. 

The lite transformer \citep{wu2020lite} proposes a Long Short Range Attention (LSRA), which preserves the capacity of self-attention layers to model relationships in the input sequence. LSRA removes the dimensionality reduction of the bottleneck configuration observed in the feed-forward modules surrounding the self-attention layers in the transformer's original configuration. Then, it replaces the self-attention layers for a dual-channel configuration. The first channel comprises self-attention heads for long-range modeling. The second channel is a convolutional module for short-range modeling. The two channels do the processing in parallel, each with a chunk of the input. Then, the results are concatenated with a feed-forward module. 

The number of parameters in LSRA remains the same, but there is no context loss like the loss that the dimensionality reduction produces in the original transformer architecture. No heads in the self-attention module are reduced, while the context reduction is removed by flattening the input feed-forward module. Similarly, the convolutions in the second channel are faster and less affected by sequence length. Efficiency improvements can go up to 2 times. Moreover, using 8-bit quantization and pruning, the model can be up to 18 times smaller, which is ideal for smartphones and edge devices \citep{wu2020lite}.

Additional architectural optimizations include improvements to recurrent layers \citep{moumen2023stabilising,ravanelli2018light}. Though transformers are widely used in speech processing, recurrent networks continue to find applications in offline and online speech processing \citep{chung2018speech2vec,le2019senones,moumen2023stabilising,ravanelli2018light,wang2020transformer,zhang2016highway}. For example, the DeepSpeech architecture uses convolutional and recurrent layers for ASR \citep{amodei2016deep}; therefore, recurrent networks coexist with transformers in speech processing \citep{moumen2023stabilising}.

Light gated recurrent units (LiGRU) were proposed for ASR \citep{ravanelli2018light}. They reduce by 30\% the training time per epoch when compared against standard GRUs. LiGRUs suppress one gate, the reset gate, reducing the number of parameters in the recurrent unit. LiGRUs also replace the tanh activation function for a ReLU function. However, no implementation was released to the community, hindering its adoption. Existing implementations of LiGRUs use no recurrent optimization, making them slower than optimized LSTM implementations. Another serious limitation is the exploding gradient problem, given the unbound nature of ReLU activations. Even though at first it used batch normalization to deal with exploding gradients, large sequences in speech processing make the problem resurge \citep{moumen2023stabilising}.

An improved version of LiGRUs was recently proposed \citep{moumen2023stabilising}: Stabilized LiGRU (SliGRU), which adds layer normalization to the recurrent units to solve the exploding gradients. Besides, an optimized implementation using Pytorch and CUDA makes proposed units up to five times faster than LSTMs, with performance improvements in ASR. SliGRUs are also faster than a version using sine activations for dealing with exploding gradients.

\section{Towards finetuning efficiency} \label{sec:finetuning_efficiency}

Parameter efficient transfer learning (PETL) aims at benefiting from pretrained representations and fine-tuning models to perform downstream tasks \citep{lialin2023scaling,wang2023read}. Recent self-supervised approaches use large transformer architectures, which are fine-tuned for specific tasks after a pretraining phase, transferring learning from a pretraining objective to a downstream task. As the size of the transformer models increases, fine-tuning the whole model becomes costly because of the increase in the number of trainable parameters, and for several downstream tasks, there must be one fine-tuned model per task \citep{wang2023read,zhang2023adaptive}

To address this limitation, several PETL approaches have been proposed to reduce the computational cost of fine-tuning large transformer models. PETL approaches include reparameterization, additive methods, soft prompts, and partial tuning \citep{wang2023read}.

In partial tuning, only certain layers are trained, or only certain parameters of the model, such as the biases, are trained during the fine-tuning phase \citep{zaken2022bitfit}. Prompting approaches concatenate trainable input vectors to the model input, training them during fine-tuning to modify the transformer output, or alter the whole input embeddings, which represents a learnable prompting scheme \citep{lester2021power,li2021prefix}. Additive methods insert small neural components into the backbone transformer, training only the components during fine-tuning. Components can be layers inside a transformer or external components in a side-tuning or ladder side-tuning configuration \citep{sung2022lst,wang2023read,zhang2020side}. Finally, low-rank components alongside self-attention layers are part of reparameterization approaches \citep{hu2021lora,karimi2021compacter,zhang2023adaptive}.

Low-Rank Adaptation (LoRA) is a reparameterization approach \citep{hu2021lora}. LoRA adds low-rank decomposition matrices to each transformer layer and trains the matrices during fine-tuning, keeping the weights of the pretrained model frozen. LoRA draws inspiration from the fact that learned transformer models have a low intrinsic dimensionality, suggesting that learning during fine-tuning also happens in a low dimensionality. It is the first approach to add low-rank restrictions for downstream tasks only. 

Advantages of LoRA include the same or better results than the fine-tuned baseline, reduction of trainable parameters up to 10 thousand times for large models such as GPT3 \citep{brown2020language}, which has 175B parameters, reduction of GPU RAM usage of up to three times, and no latency during inference. As far as the limitations of LoRA are concerned, low-rank matrices are added to the model to avoid latency; however, this addition limits parallelization when using a batch with different downstream tasks. Likewise, heuristics dictate the way low-rank configuration is selected, motivating the need for more principled ways of setting the configuration. Moreover, LoRA does not prioritize the parameters that have the highest impact on the model, which makes the fine-tuning suboptimal. It also adds low-rank matrices to the self-attention layers only, while low-rank matrices for the feed-forward layers have a higher impact \citep{hu2021lora,zhang2023adaptive}. 

As an alternative, Adaptive Low-Rank Adaptation, or AdaLoRA \citep{zhang2023adaptive}, creates low-rank matrices for both self-attention layers and feed-forward layers in the transformer model. AdaLoRA dynamically adapts the low-rank matrices during the fine-tuning process. While using singular value decomposition allows us to identify the most important ranks during fine-tuning, its computational cost is too high. Instead, AdaLoRA uses three low-rank matrices $PVQ$, where $P$ and $Q$ store the singular dimension vectors, while $V$ stores the singular values. $V$ is diagonal; thus, it can be stored in a vector. $P$ and $Q$ are orthogonal so that low-rank dimensions remain independent from each other. After each gradient descent, an importance metric allows us to determine whether a singular value remains important for the model output. Should a singular value be evaluated as not important, its dimensions are not deleted. Instead, the singular value is replaced by a zero in the $V$ matrix. This pruning allows the singular dimensions to be reactivated later if their importance increases again. 

Though the aforementioned PETL approaches reduce the number of parameters trained during fine-tuning, memory consumption remains high. A similar issue happens with energy consumption. Similarly, side-tuning fails to account for intermediate layer results, which are crucial for representation learning, taking only intermediate activations. Ladder side-tuning inserts a transformer component to consider intermediate layer results during fine-tuning, but the inserted transformer requires an additional pretraining phase \citep{wang2023read}.

To address these issues, Recurrent adaptation of large transformers (READ) proposes an additive neural component for fine-tuning \citep{wang2023read}. The neural component comprises a joiner network and a recurrent module without any additional attention mechanisms. The joiner takes as input the output hidden states from all the transformer layers, processing them with a feed-forward network. Then, the recurrent module iteratively processes the joiner output. The final model output is the addition of the recurrent component output and the transformer output.

Tests were performed using the T5 architecture, an encoder -- decoder transformer model \citep{raffel2020exploring}. The dataset used in tests is GLUE \citep{wang2018glue}, which includes several NLP downstream tasks such as natural language inference, sentiment classification, paraphrase detection, and linguistic acceptability. Limitations of READ include its behavior in low data configurations. Preliminary tests show READ needs more epochs than existing methods when fine-tuning for small datasets \citep{wang2023read}.

In contrast, the advantages of READ include a 50\% reduction of RAM and 86\% of GPU energy consumption against fine-tuning with the whole model. Also, the number of additional parameters increases in log-linear form against the transformer's size. Memory usage and energy consumption improvements are also observed when comparing READ against existing PETL methods. Another advantage is multitasking. As the neural component added for fine-tuning is small, it is possible to add multiple components, each one for a task in a multitask setup, without the need to fine-tune separately for each task \citep{wang2023read}.

\section{Towards data efficiency for model pretraining} \label{sec:data_efficiency}

SSL models enable data efficiency when finetuning for downstream speech tasks, as pretrained speech representations can facilitate finetuning with small datasets \citep{ericsson2022self}, motivating research for data-efficient finetuning with very low datasets. But there is also the possibility to improve data efficiency for pretraining SSL models. Recent work in data efficiency for pretraining includes dataset distillation \citep{maekawa2023dataset,wang2018dataset}, trimming of speech sequences \citep{gaol2023match}, contrastive learning for generative models \citep{shi2020relating}, and hierarchical pretraining \citep{reed2022self}.

Hierarchical pretraining proposes a multiphase approach. First, the model is trained with a general dataset. Then, the pretraining continues with a domain-specific dataset. In the third phase, the model is pretrained with the target dataset before applying the supervised finetuning with the target dataset. In comparison, generalist pretraining uses a large dataset combining many domains in the input samples, which enables the creation of a repository of models. Specialist pretraining uses a domain-specific dataset to learn the model. In both generalist and specialist pretraining, a final phase of supervised finetuning is realized for calculating model results. But in certain cases, generalist pretraining underperforms models trained end-to-end with domain-specific data. Hierarchical pretraining addresses this issue. It can also improve training convergence, making it up to 80 times faster \citep{reed2022self}.

Generative models using multimodal data focused on the commonality between related samples of different modalities - for example, matching images to their text description. Recent work \citep{shi2020relating} extended the relationship to include non-related samples in the loss calculation. Non-related samples are negative examples in a contrastive learning configuration. By doing so, the multimodal variational autoencoders improve their performance over existing models, optimizing the amount of data necessary for training.  

Based on previous results from federated learning for speech, short sequences effectively reduce the computational costs of speech models \citep{gao2022federated}. Therefore, a systematic analysis of short sequences enables their evaluation in speech processing tasks, trimming speech sequences to 1, 5, 10, and 15 seconds. Several downstream tasks are analyzed, performing model training with edge platforms like Raspberry Pi and Nvidia Jetson Xavier. Experimental results show the computational cost reduction that a one-line code change represents, with acceptable results in downstream tasks, especially for downstream applications requiring short speech sequences \citep{gaol2023match}. 

Data distillation focuses on data efficiency for training neural models. This method generates synthetic samples from a dataset to train models in a few training steps \citep{maekawa2023dataset}. Synthetic samples enable a form of dataset compression. For example, one synthetic image per class in the MNIST dataset enables training a model with a fixed network initialization in a few iterations \citep{wang2018dataset}.

\section{Conclusions} \label{sec:conclusions}

In sum, this paper presented an overview of self-supervised representation learning for speech processing, including contrastive, predictive, and multilingual models. Then, we introduced efficiency-oriented approaches like optimization of existing models, neural architecture modifications, finetuning improvements, and data efficiency.  

There are, however, several future research directions for optimizing self-supervised representation learning in speech processing. First, despite efforts to improve model efficiency \citep{chen2023reducing,parcollet2023efficiency}, more work should be done to reduce the computational costs of self-supervised models. Computational costs create challenges when using these models in mobile devices, for training on very large datasets, and from an energy consumption perspective \citep{mohamed2022self,parcollet2023efficiency,wu2022sustainable}.

Secondly, given the impressive performance in several speech tasks, the learned representations seem to encode more features than simply speech. Disentangling the latent vectors could allow other applications. For instance, decomposing the latent representation to model ASR and SID simultaneously \citep{mohamed2022self,stafylakis2023extracting}. Thirdly, training datasets for natural language models use way more data than the equivalent models in speech. Natural language models could provide semantic context to representation learning, helping them to learn semantic information from language. Yet it is not clear how to integrate them in speech representation approaches \citep{mohamed2022self}. Finally, other research directions include data efficiency for model pretraining \citep{maekawa2023dataset} and capturing semantic information in speech, which has a higher abstraction level than phonetic or lexical modeling \citep{arora2022espnet,lai2021semi,mohamed2022self,seo2022integration}.

\bibliographystyle{IEEEtranSN}
\bibliography{references}

\end{document}